\documentclass[letterpaper, 10 pt, conference]{ieeeconf}  

\IEEEoverridecommandlockouts                              

\usepackage{graphics} 
\usepackage{epsfig} 
\usepackage{mathptmx} 
\usepackage{times} 
\usepackage{amsmath} 
\usepackage{amssymb}  
\usepackage{float}
\usepackage{epstopdf}
\usepackage{cite}
\usepackage{hyperref, url}
\usepackage{algpseudocode}
\usepackage{algorithm}
\usepackage{psfrag}
\usepackage{soul}
\usepackage{balance}
\makeatother
\title{\LARGE \bf
Modeling Human Strategy for Flattening Wrinkled Cloth \\Using Neural Networks}

\author{Nilay Kant$^{1}$, Ashrut Aryal$^{1}$, Rajiv Ranganathan$^{2}$, Ranjan Mukherjee$^{1}$, and Charles Owen$^{3}$ 
\thanks{$^{1}$First, second and fourth authors are with the Department of Mechanical Engineering, Michigan State University, East Lansing, Michigan, USA.}
\thanks{$^{2}$Third author is with the Department of Kinesiology, Michigan State University, East Lansing, Michigan, USA.}
\thanks{$^{3}$Fifth author is with the Department of Computer Science and Engineering, Michigan State University, East Lansing, Michigan, USA. The corresponding author is Ranjan Mukherjee: 
	{\tt\small {mukherji@egr.msu.edu}}}
 \thanks{This work was supported by the Michigan State University Strategic Partnership Grant and by the National Science Foundation, Grant No. CMMI-2326227.}
	}

\begin{document}




\maketitle

\begin{abstract}
This paper explores a novel approach to model strategies for flattening wrinkled cloth learning from humans. A human participant study was conducted where the participants were presented with various wrinkle types and tasked with flattening the cloth using the fewest actions possible. A camera and Aruco marker were used to capture images of the cloth and finger movements, respectively. The human strategies for flattening the cloth were modeled using a supervised regression neural network, where the cloth images served as input and the human actions as output. Before training the neural network, a series of image processing techniques were applied, followed by Principal Component Analysis (PCA) to extract relevant features from each image and reduce the input dimensionality. This reduction decreased the model's complexity and computational cost. The actions predicted by the neural network closely matched the actual human actions on an independent data set, demonstrating the effectiveness of neural networks in modeling human actions for flattening wrinkled cloth.
\end{abstract}


\section{Introduction}\label{sec1}
Flattening deformed or wrinkled cloth is critical in several industries, such as textiles \cite{schrimpf2012experiments, kaltsas2022review}, garments \cite{winck2009novel, koustoumpardis2004review, koustoumpardis2006intelligent}, and surgical robots \cite{seita2020deep}. Smooth, wrinkle-free fabric is essential for procedures like cutting, sewing, and packaging. Automating this task with robots presents significant challenges due to the complex and dynamic nature of fabric manipulation \cite{gershon1993strategies}. The inherent flexibility of fabrics and the wide variety of wrinkle types require high levels of precision and adaptability. Humans excel at this task, and this research aims to model human strategies for flattening wrinkled cloth to enhance automation capabilities in industries where fabric handling is essential.

Researchers have employed various approaches to address the challenge of cloth flattening using robots. For example, a heuristic-based approach in \cite{sun2014heuristic} identified clusters of wrinkles with k-means filtering on the range map, targeting the largest wrinkle and applying an appropriate force to eliminate it without creating new ones. Seita et al. \cite{seita2020deep} utilized deep imitation learning with a fabric simulator and an algorithmic supervisor, which provided paired observations and actions based on complete fabric state information. However, this approach often failed on already smooth fabrics, causing unnecessary pulls that introduced wrinkles. A dynamic method involving high-velocity actions namely,  `pick', `stretch', and `fling' using a dual-arm robot was proposed in \cite{ha2022flingbot}, where a self-supervised learning framework that learns from visual observations was employed. A two-phase algorithm for automating the unfolding and flattening of laundry using interactive perception was presented in \cite{willimon2011model}. This paper presents an approach to flattening wrinkled cloth by learning from human behavior, a method that contrasts with existing techniques.

Understanding human behavior is crucial for developing effective human-robot interactions and enhancing the safety and efficiency of such systems \cite{yang1997human, psarakis2022fostering}. The need for robots to be highly aware of their surroundings and human counterparts increases with more frequent interactions \cite{jahanmahin2022human}. For instance, robots can be trained with statistical models of human behavior to make decisions aligned with individual human styles \cite{nikolaidis2015efficient}. Learning from human behavior has proven beneficial in various fields. In autonomous driving, Olier et al. \cite{olier2017dynamic} developed a method that enables vehicles to replicate human driving behaviors through deep learning and Bayesian filtering, enhancing autonomous monitoring. Kuperwajs et al. \cite{kuperwajs2023using} used deep neural networks to improve cognitive models of human planning by analyzing gameplay patterns, refining decision-making models. Wang et al. \cite{wang2018human} introduced an RGB-based architecture combining CNNs and LSTM units with a temporal-wise attention model to efficiently recognize human actions.

Motivated by the advantages of machines replicating human behaviors, we examine how humans flatten wrinkled cloth by collecting data on finger placement, pull direction, and pull length from human participants, relative to the cloth's wrinkled state. This data is used to train a regression neural network, which effectively models these human actions and can be applied to robotic automation for flattening wrinkled cloth, in sewing industry, for example.




\section{Methods}\label{sec2}
\begin{figure}[b!]
\centering
\includegraphics[width=1.0\linewidth]{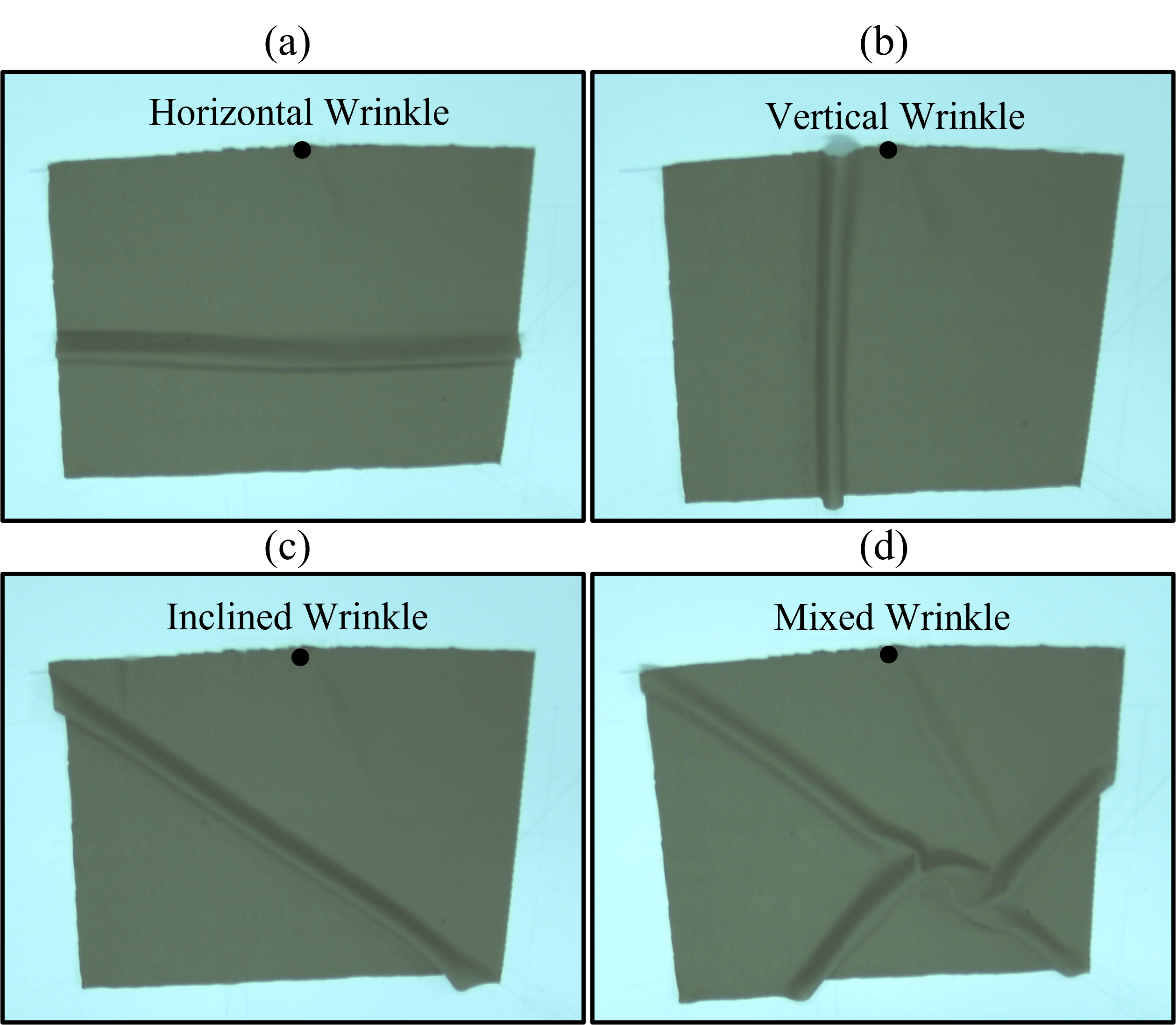}
\caption{Four wrinkle types used in experiments: (a) horizontal, (b) vertical, (c) inclined and (d) mixed. Black dot shows location where the cloth is pinned to the table.} 
\label{Fig1}
\end{figure}

\subsection{Participants}\label{sec2a}
Ten college-aged individuals in engineering, with no known motor impairments volunteered for the experiment. Participants provided written informed consent, and the experimental procedures were approved by the Institutional Review Board at Michigan State University.
\begin{figure}[t!]
\centering
\includegraphics[width=0.91\linewidth]{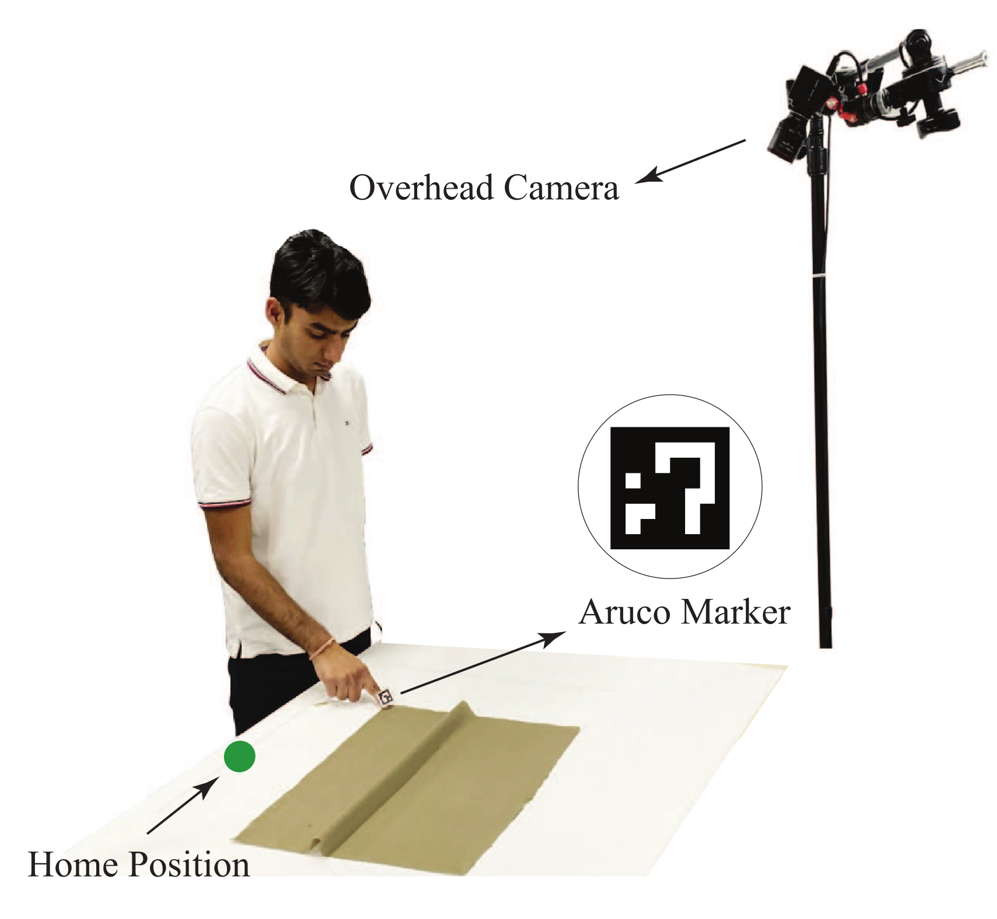}
\caption{Experimental setup for data collection} 
\label{Fig2}
\end{figure}

\subsection{Task Description}\label{sec2b}
\begin{figure*}[t!]
\centering
\includegraphics[width=0.87\textwidth]{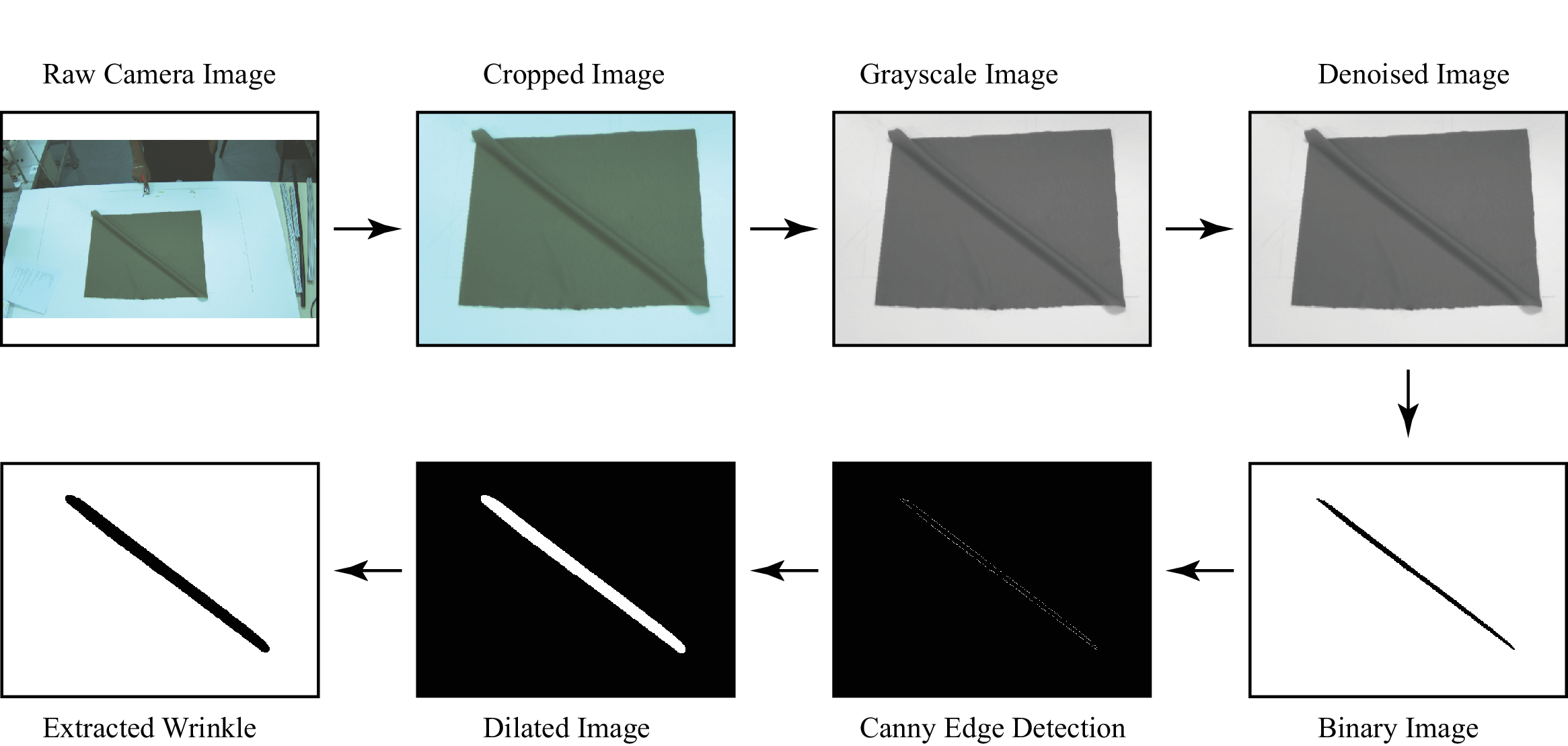}
\caption{Image processing steps utilized for extracting wrinkle features.}
\label{Fig3}
\end{figure*}
The objective was to flatten a rectangular, uniformly-colored, wrinkled cloth placed on a flat table. To ensure that the cloth did not translate freely over the table, the midpoint of the top edge of the cloth was pinned to the table. Four distinct wrinkle patterns were employed in the study: vertical, horizontal, inclined, and mixed - see Fig.\ref{Fig1}. The point where the cloth was pinned to the table is indicated by a black dot. Each wrinkle type was presented five times to each participant, resulting in twenty trials per participant. Wrinkle patterns were generated before each trial by laying the cloth on a rod placed on the table and then removing the rod; this allowed us to generate the different types of wrinkles consistently across different trials and participants.\

The objective of the study was to examine the strategy used by humans for flattening wrinkled cloth. Participants were encouraged to accomplish flattening of the cloth with the minimum number of iterations. When a wrinkle pattern was presented, participants were directed to use their index finger to pull the cloth from any point on an edge of the cloth along a straight line. Prior to each pull, participants were instructed to position their index fingers at a designated home position on the table - see Fig.\ref{Fig2}. The sequences of actions comprised of lifting the finger from the home position, translating it to the desired position on the edge of the cloth, executing the pull, and returning the finger to the home position, constituted one iteration. Participants were permitted to proceed at their own comfortable pace and to pause in the home position to strategize their next iteration. The iterative process was continued till the cloth was in a state of near flatness (as determined visually by the experimenter).  It should be noted that each participant was given a practice trial in the beginning to get familiar with the process. Additionally, the order in which the wrinkle patterns were presented to the participants was randomized.

\subsection{Data Acquisition}\label{sec2c}
The experimental setup for the study is shown in Fig.\ref{Fig2}, and is comprised of an overhead camera and an Aruco marker on the participant's finger. We employed a 12 MP Arducam IMX477 camera system and an Arducam 2.8-12mm Varifocal C-Mount lens at 147cm distance from the center of the cloth. The camera was configured to capture images at a resolution of $3840 \times 2160$ pixels, resulting in a 8MP output. The lens was set to a $12mm$ focal length for a $37.7$ degree horizontal field of view. Image acquisition was conducted using a local PC equipped with the OpenCV Python library. To track the motion of the index finger, Aruco markers \cite{garrido2014automatic} were used. 
Real-time motion tracking was achieved using the OpenCV Aruco library. The camera system was used to simultaneously sense both the state of the cloth and finger movements. The collected data for each trial included image frames along with the corresponding 3D-coordinates of the Aruco marker. This data was utilized to extract the image of the wrinkled cloth prior to each iteration and the attributes of the iteration, which include the Cartesian coordinates of the location where the finger is placed for the pull operation, and the direction and length of the pull performed by the participant. 

\section{Image Processing for Wrinkle Extraction}\label{sec3}

Our aim was to establish a connection between the state of the wrinkled cloth and the corresponding action (attributes of iteration) for its removal by a human participant. It was postulated that variables such as orientation, location, shape, and height of the wrinkles would impact the decision-making process involved in pulling the cloth. Utilizing an overhead camera setup, images of the state of the cloth were captured prior to each iteration. These images, however, contained superfluous information such as the region outside the perimeter of the cloth, which does not contain relevant information for decision-making. Additionally, we reasoned that while wrinkles play a significant role in decision making, finer details such as thread patterns are less influential. Therefore, our methodology focused on extracting solely the wrinkle features from the camera image frame captured prior to each iteration.

To extract wrinkles from a raw camera image, a series of image processing techniques were employed. The raw image captured by the camera, comprising of RGB channels, was first cropped to isolate the cloth portion while retaining background elements on the table and scaled to a size of 100 x 100 pixels. The image was then converted to monochrome. To reduce image noise, a non-local means-based filtering technique \cite{buades2005non} was applied. Wrinkled cloth exhibits distinct edges when viewed from an overhead camera. To separate these edge-like regions from the cloth background, the image was binarized using Otsu's adaptive thresholding method \cite{otsu1975threshold}. Notably, the obtained threshold value was scaled down (60 \%) for consistent background removal across all trials. Following the separation of wrinkle features from the background, Canny's edge detection algorithm \cite{canny1986computational} was utilized to highlight the wrinkle boundaries. Next, the morphological operation of dilation \cite{haralick1992computer} with a disk structuring element was used to refine and emphasize the wrinkle feature. Finally, the dilated image was inverted to achieve a clean extraction of wrinkles against a white background. A schematic of the image processing\footnote{ All image processing operations were performed using the computer vision toolbox in Matlab.} steps for an inclined wrinkle is presented in Fig. \ref{Fig3}.\

\section{Learning Human Strategy}\label{sec4}

\subsection{Consistency in Human Strategy}\label{sec4a}

\begin{figure}[b!]
\centering\includegraphics[width=1.0\linewidth]{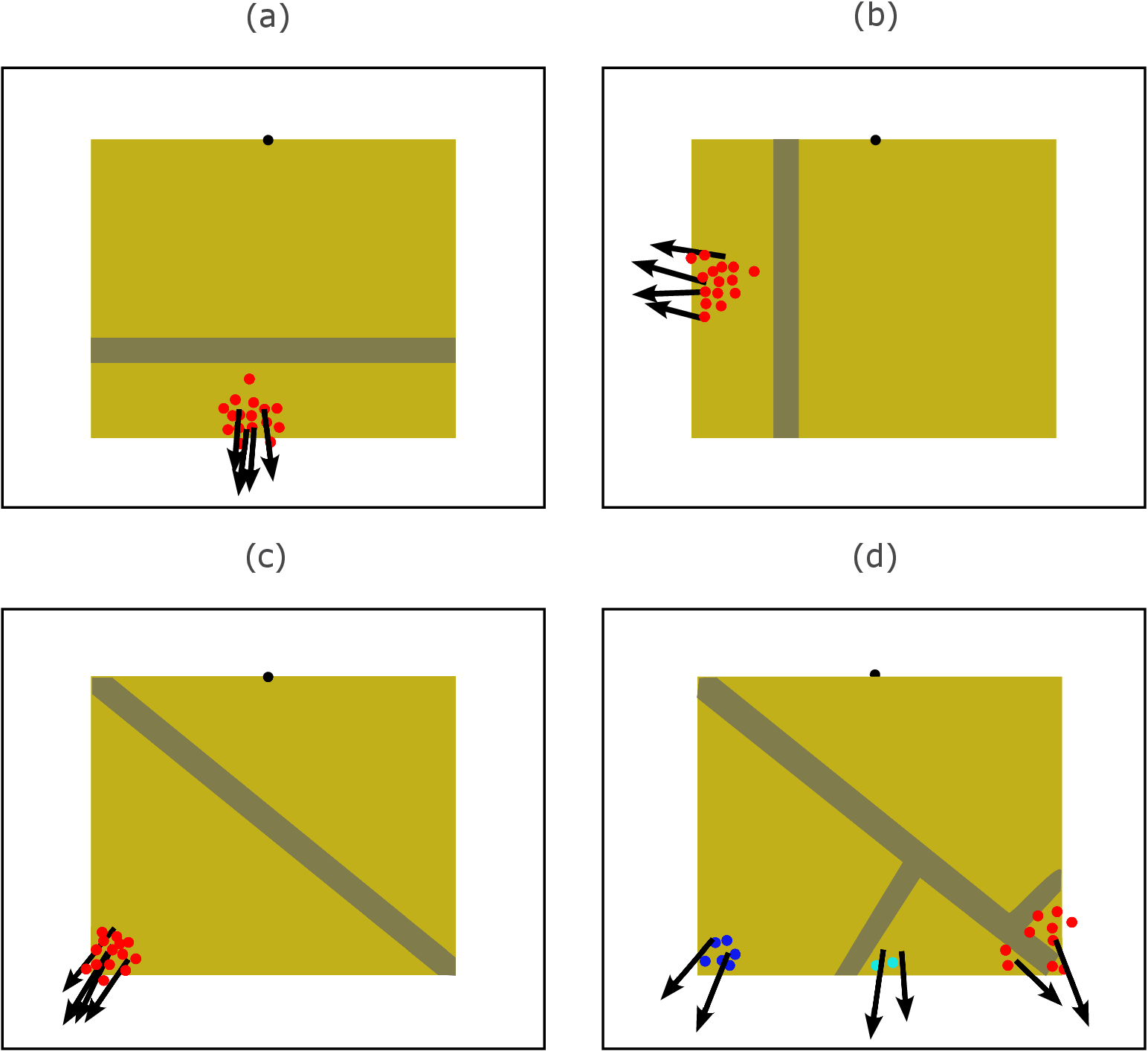}
\caption{Schematic illustrating human strategy for wrinkle removal from (a) horizontal, (b) vertical, (c) inclined, and (d) mixed wrinkle types. Red dots denote the location of finger placement in the first three sub-plots where there is only one dominant region of finger placement. In the fourth sub-plot, different colors represent each of the three major regions of finger placement. Black arrows depict the length and direction of pull operation.}
\label{Fig4a}
\end{figure}

The human participant data for the four wrinkle types is shown in Fig.\ref{Fig4a}. The red dots denote the position of finger placement prior to a pull operation and the black arrows\footnote{All the arrows were concentrated in a small region and therefore a few of them are shown to avoid over-crowding} depict the pull length and direction. It can be seen in Fig.\ref{Fig4} that the human strategy is consistent in terms of position, direction, and length of pull for three of the four wrinkle types, namely, horizontal, vertical and inclined. The human strategy for the mixed wrinkle type - see Fig.\ref{Fig4a}(d), was inconsistent. The initial pull locations were clustered in three different regions at the bottom edge of the cloth - the two corners and the midpoint. These strategies were also equally effective in flattening the cloth, as participants, on average, flattened the cloth in two iterations. This observation suggests the existence of multiple optimal solutions for flattening a mixed wrinkle. Due to the absence of a consistent human strategy, we excluded the mixed wrinkle category from our current analysis.

\subsection{Image Dimensionality Reduction}\label{sec4b}

\begin{figure}[t!]
\centering
\includegraphics[width = 0.82\linewidth]{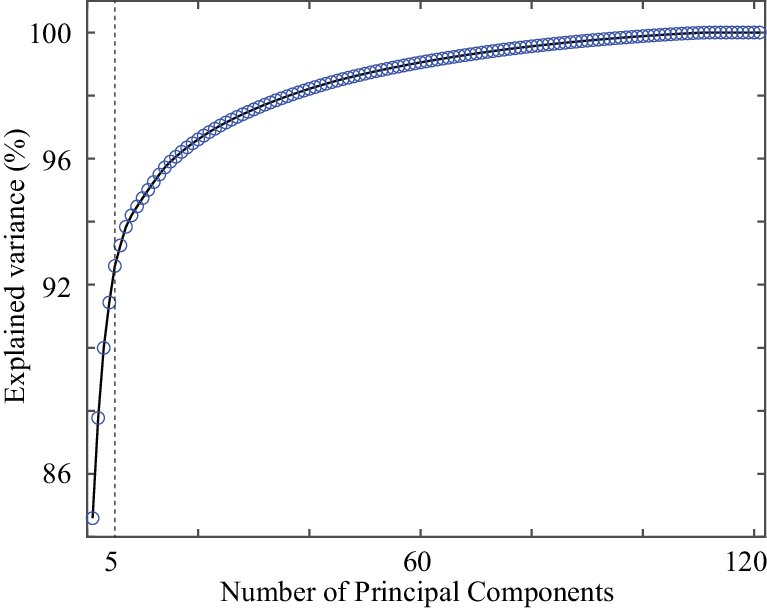}
\caption{Plot of explained variance of the input image data.} 
\label{Fig4}
\end{figure}
Our objective is to learn the human strategy for manipulating wrinkled cloth based on its visual representation. To this end, we utilize a regression neural network framework for supervised learning. It is important to note that we assume human actions depend solely on the current cloth state and are not influenced by past cloth states or actions. The images processed in Section \ref{sec3} are of dimensions of $100 \times 100$ pixels. Our dataset contains 112 processed images containing three types of wrinkle patterns: vertical, horizontal, and inclined. We also included $10$ images of completely flat cloth, which would require no pulling action. This resulted in an input dataset of dimension $122 \times 10000$, where each row of the array represents an image. The corresponding human action to each of these $122$ trials is characterized by  four parameters: the coordinates ($x$ and $y$) of finger placement for cloth manipulation, the length of the pull ($d$), and the direction of the pull ($\theta$). The human actions linked with flat cloth was set as a vector with four zero entries. This resulted in an output dataset of dimension  $122 \times 4$.\

\begin{figure}[t!]
\centering
\includegraphics[width = 0.91\linewidth]{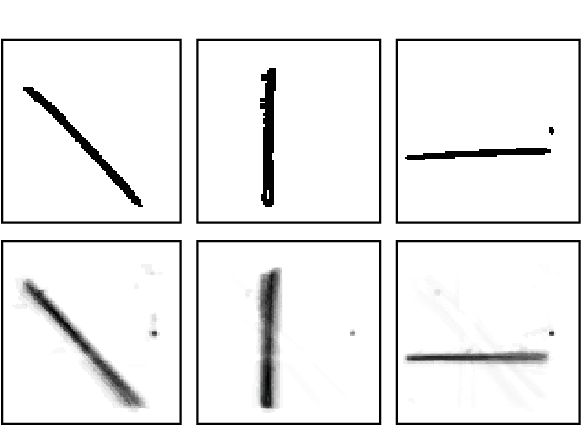}
\caption{Comparison of original and reconstructed images. Top row shows inclined, vertical, and horizontal wrinkle. Bottom row shows the corresponding images reconstructed with only five principal components.} 
\label{Fig5}
\end{figure}
In theory, a regression neural network could be directly trained on this data. However, our training dataset is constrained due to the limited number of participants. Given the high dimensionality of the input data, relying solely on the neural network to extract relevant features would significantly increase the complexity of the network, thereby expanding the number of parameters to be learned. This may not be feasible given the limitations imposed by the size of the dataset. To determine the number of relevant features in our input dataset, we employ principal component analysis (PCA) \cite{abdi2010principal}. The cumulative explained variance plot, shown in Fig.\ref{Fig4}, indicates that five principal components contribute to over $92\%$ variation within the images. Therefore, we opted for 5 principal components and projected our original dataset, which had a dimension of $10000$ (pixels), to a significantly reduced dimension of $5$. To visualize the effectiveness of this dimensionality reduction, plots of three sample wrinkle types are presented in the first row of Fig.\ref{Fig5}; the second row displays the same images reconstructed using only the five principal components. It can be observed that the five principal components distinctly retain the wrinkle attributes.

The input dataset was reduced from $122 \times 10000$ to $122 \times 5$ using PCA prior to training. To address variations in scale and units among output variables, z-score standardization was applied. Before training, a randomization procedure was implemented on both input and output data to shuffle samples, thereby mitigating potential biases inherent in the dataset. Subsequently, the dataset was partitioned into training and validation sets at a ratio of 75\% to 25\%.
 
\begin{figure}[t!]
\centering
\includegraphics[width = 1.0\linewidth]{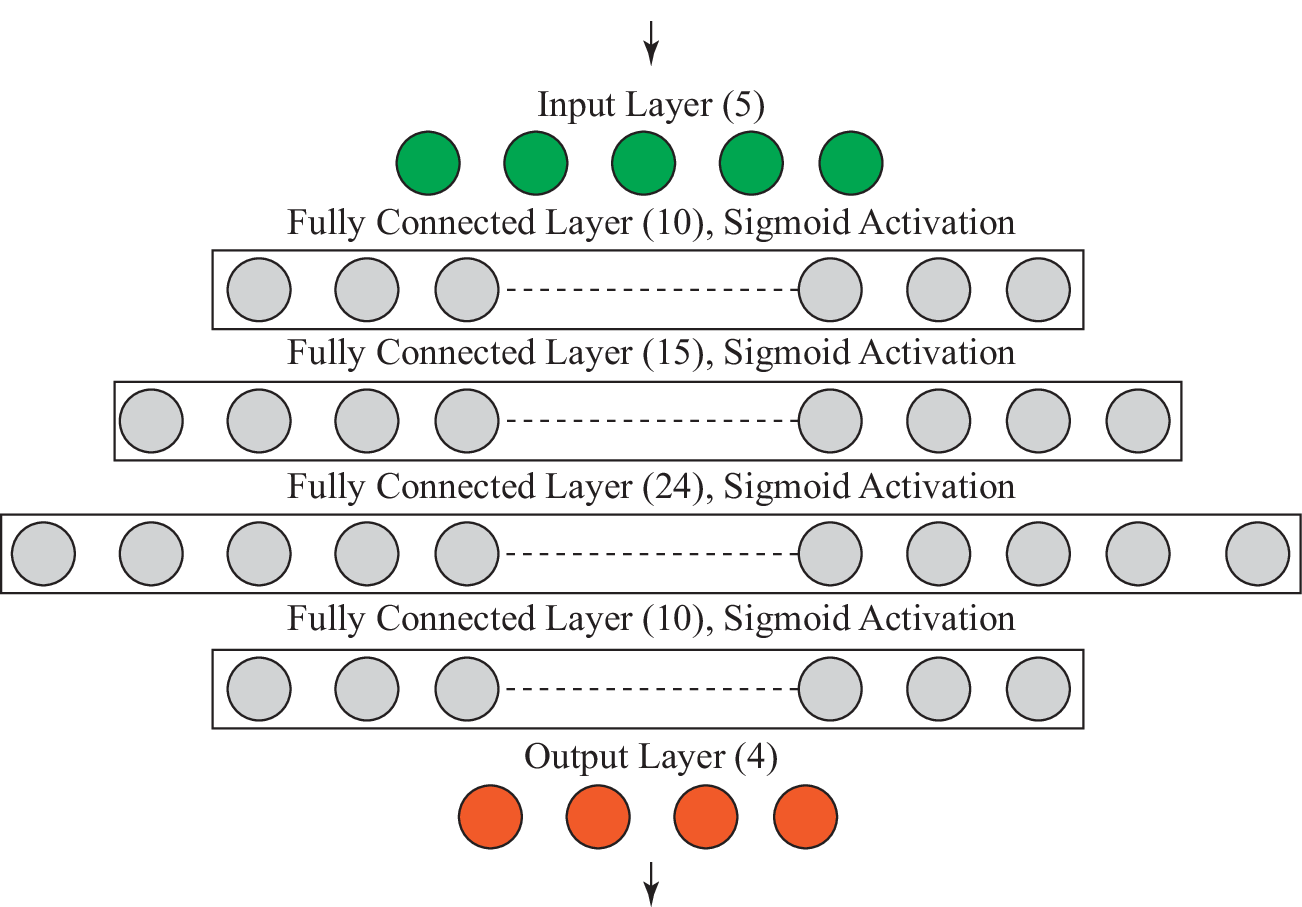}
\caption{Neural network architecture.} 
\label{Fig6}
\end{figure}

The neural network architecture, shown in Fig.\ref{Fig6} comprised of four fully connected hidden layers. The initial layer featured 10 neurons, followed by a layer with 15, 24, and 10 neurons, respectively. The sigmoid activation function was incorporated after each fully connected layer to capture nonlinear relationships between input and output data. The output layer, consisting of 4 neurons, was also fully connected, followed by a regression layer to facilitate continuous valued predictions, aligning with the output dimensionality.
\subsection{Neural Network Description and Training}\label{sec3d}

Three optimization algorithms namely, stochastic gradient descent, RMSprop, and Adam were evaluated in training the neural network. Among these, the Adam optimizer \cite{kingma2014adam} exhibited superior training performance. This can be attributed to the observation that the gradient exhibited fluctuations during the training process. Consequently, the Adam optimizer, known for its robustness to noisy gradients, was selected for learning the neural network weights. The training data was shuffled every epoch. The gradient decay factor was set to 0.9, and the squared gradient decay factor to 0.999, ensuring smooth convergence and stability during training. The initial learning rate was set to 0.01, with no predefined learning rate schedule, but a drop factor of 0.1 is applied every 10 epochs to adaptively adjust the learning rate. $L_2$ regularization with a weighting of $1.0 \times 10^{-4}$ was employed to reduce over fitting. Training was carried out for a maximum of 100 epochs, utilizing mini-batches of size 12. Initially, the root mean squared error was approximately 2.8. Around the 90th epoch, the average root mean squared error stabilized within a close range of 0.8.\

\section{Results}\label{sec4}

\begin{figure}[t!]
\centering
\includegraphics[width = 1.0\linewidth]{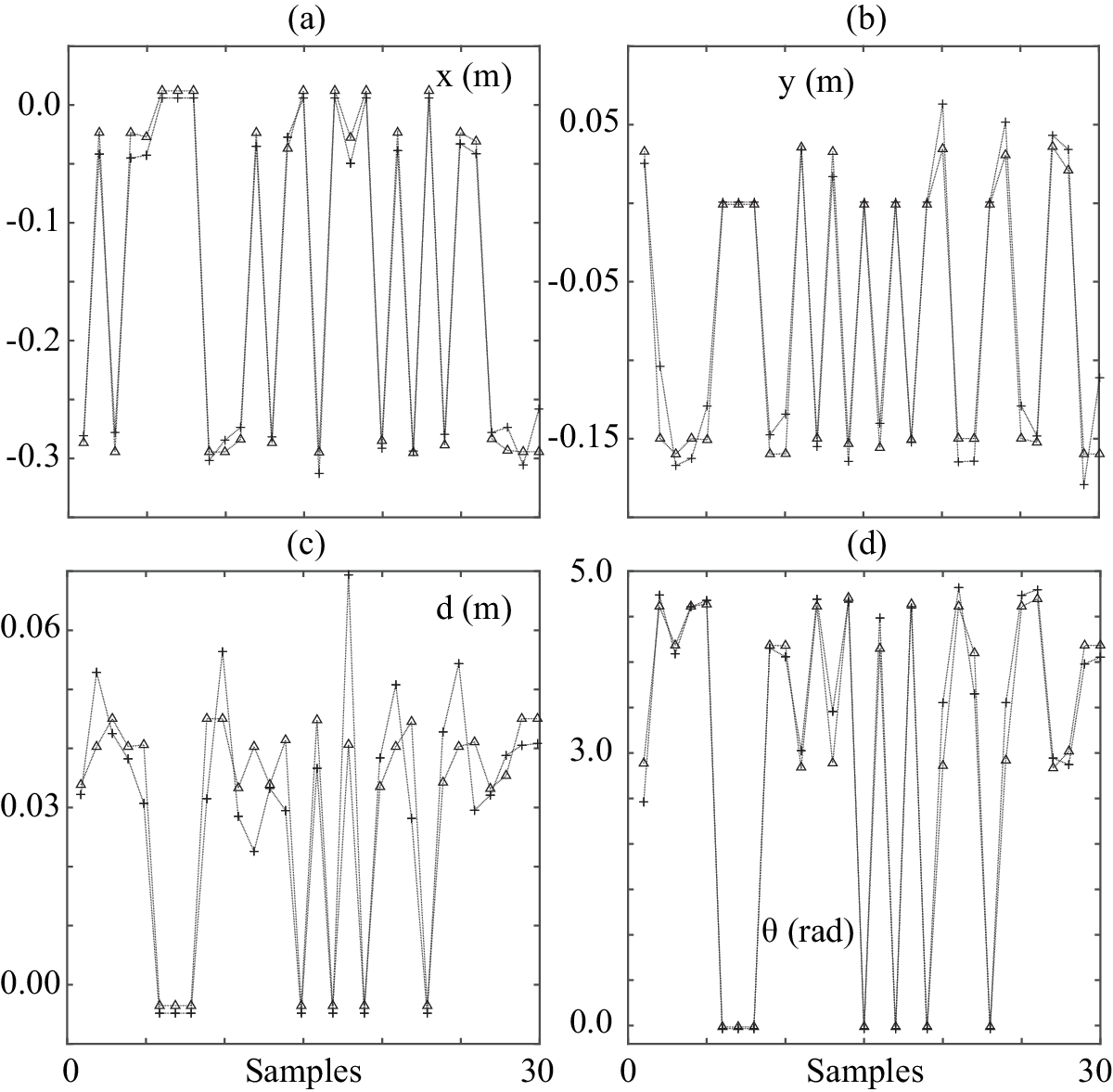}
\caption{Plot showing neural network performance in learning human strategy. The symbols $+$ and $\triangle$ denote the actual and predicted human actions, respectively.} 
\label{Fig7}
\end{figure}

After training the neural network, its performance was assessed using the validation dataset. The actual and predicted plots of the four human actions are displayed in Fig.\ref{Fig7}. Figs.\ref{Fig7} (a) and (b) show the actual and predicted $x$ and $y$ coordinates of finger placement before each pull, while Figs.\ref{Fig7} (c) and (d) show the length of pull and direction of pull for both actual data and those predicted by the neural network. The root mean square (RMS) errors of the neural network prediction for $x$ and $y$ coordinates, pull length (d), and pull angle ($\theta$) are provided in Table \ref{tab1}.

\begin{table}[b!]
\centering
\caption{RMS Prediction Error of Human Actions}
\renewcommand{\arraystretch}{1.5} 
\setlength{\tabcolsep}{15pt} 
\begin{tabular}{ |c|c|c|c| } 
 \hline
 x (m) & y (m) & d (m) & $\theta$ (rad) \\ 
 \hline
 0.013 & 0.017 & 0.008 & 0.3 \\ 
 \hline
\end{tabular}\label{tab1}
\end{table}

\begin{figure}[b!]
\centering
\includegraphics[width = 0.91\linewidth]{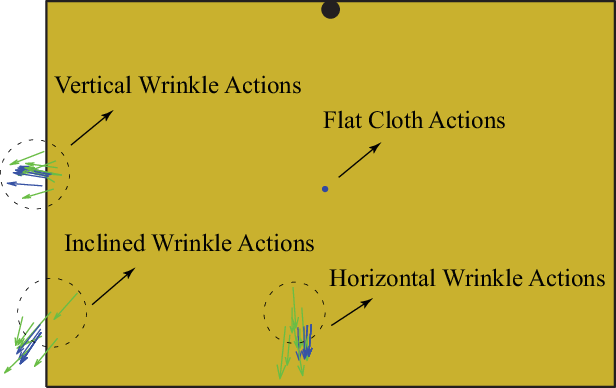}
\caption{Real and predicted human actions in relation to flattening wrinkled cloth. The green  and blue arrows represent the actual and predicted human actions, respectively.} 
\label{Fig8}
\end{figure}
 
For better visualization, a schematic representing the human actions is presented in Fig. \ref{Fig8}. The yellow box represent the cloth when it is perfectly flat. The green arrow lines depict the actual human actions, while blue arrow lines signify actions predicted by the neural network. The starting points of the arrow lines indicate the initial finger placement before pulling while its length denote the pull length. Three clusters of arrow lines are observable: the upper left cluster correspond to vertical wrinkles, the lower corner cluster correspond to inclined wrinkles, and the bottom-right cluster correspond to horizontal wrinkles. It can be observed that the pull direction is approximately perpendicular to the orientation of the wrinkles. Also, for cloth in a flat state, the human actions are clustered near the origin with the length of the arrows almost equal to zero. This is shown by a blue dot.\

The actual and predicted human actions are closely clustered together for each of the three wrinkle types. Note that some arrows in Fig.\ref{Fig8} may appear either within or outside the cloth even though participants pulled the cloth from its edges; however this discrepancy in the visualization is due to the fact that the edges of the cloth shift when wrinkles are formed and this results in a visible cloth area that is smaller than depicted. Additionally, since the cloth was fixed at one point, it occasionally underwent slight rotation when wrinkles were formed prior to data collection, rather than remaining perfectly horizontal. 
These variations in cloth state, along with corresponding variations in human action, are inherently included in the neural network modeling. As evident from Figs. \ref{Fig7} and \ref{Fig8}, the neural network closely models and predicts human strategies for flattening wrinkled cloth.


\section{Discussion}\label{sec5}

Flattening wrinkled cloth is commonly done by humans, yet the decision-making process underlying this task remains unclear. To address this, we conducted a human participant study where participants were tasked with flattening pre-formed wrinkled cloth, considering four distinct wrinkle patterns: horizontal, vertical, inclined, and mixed. Participants were instructed to pull the cloth from its edges along a straight line, while their finger motion trajectories and cloth images were recorded using a camera. Four variables, namely, Cartesian coordinates of finger placement before pulling, pull length, and pull direction, constituted human action variables. Our objective was to map input wrinkled cloth images to these action variables using a supervised regression neural network model. Image processing and PCA were utilized for feature extraction due to the limited number of human participants. The network was then trained and validated using human participant data, demonstrating promising results in predicting human actions.

A limitation of this study is that it modeled human actions for only a single wrinkle on cloth due to the challenges and costs of acquiring data from human trials. To achieve good prediction performance, we kept the network complexity low, as increasing it would require more data. Additionally, for mixed wrinkle patterns, we observed three variations in human strategy, each removing wrinkles with the same number of iterations, leading us to exclude this type from our analysis. This suggests that completely learning the human approach to any wrinkle type may be nondeterministic. However, even simple wrinkle patterns provide valuable insights into human cloth-flattening approaches, which can inform the design of generalized, human-inspired algorithms. This is the focus of our current research. Future work will aim to use the network to assist a robotic arm in real-time cloth flattening during sewing.
\balance
\bibliographystyle{plain}
\bibliography{ref}
\end{document}